\documentclass[a4paper,11pt]{llncs}
\usepackage{euscript}
\usepackage{microtype}
\usepackage{mathptmx}
\usepackage[small,compact]{titlesec}
\usepackage{amssymb}
\usepackage{graphicx}
\usepackage{xspace}
\usepackage{color}
\usepackage{paralist} 
\usepackage{subfigure}
\usepackage{hyperref}
\usepackage{multitoc}
\usepackage{setspace}
\renewcommand{\c}[1]{\ensuremath{\EuScript{#1}}}
\renewcommand{\b}[1]{\ensuremath{\mathbb{#1}}}

\title{Generating a Diverse Set of High-Quality Clusterings\thanks{This research was partially supported by NSF award CCF-0953066 and a subaward to the University of Utah under NSF award 0937060 to the Computing Research Association.}}
\author{Jeff M. Phillips \and Parasaran Raman \and Suresh Venkatasubramanian}
\institute{School of Computing, University of Utah \\ \textsl{\{jeffp,praman,suresh\}@cs.utah.edu}}

\begin{document}

\maketitle

\begin{abstract}
We provide a new framework for generating multiple good quality partitions (clusterings) of a single data set.  Our approach decomposes this problem into two components, generating many high-quality partitions, and then grouping these partitions to obtain $k$ representatives. The decomposition makes the approach extremely modular and allows us to optimize various criteria that control the choice of representative partitions.  
\end{abstract}

\section{Introduction}
Clustering is a critical tool used to understand the structure of a data set. There are many ways in which one might partition a data set into representative clusters, and this is demonstrated by the huge variety of different algorithms for clustering~\cite{Bonner,Michaud1997135,springerlink:10.1009,Jain:2010:DCY:1755267.1755654,Xu:2009:CLU:1483087,citeulike:4065272,Tan:2005:IDM:1095618,Das:2009:MC:1592938,Jain:1999:DCR:331499.331504}.

Each clustering method identifies different kinds of structure in data, reflecting different desires of the end user. Thus, a key exploratory tool is identifying a diverse and meaningful collection of partitions of a data set, in the hope that these distinct partitions will yield different insights about the underlying data. 

\paragraph{Problem specification.}
The input to our problem is a single data set $X$. The output is a set of $k$ partitions of $X$.  A \emph{partition} of $X$ is a set of subsets $\c{X}_i = \{X_{i,1}, X_{i,2}, \ldots, X_{i,s}\}$ where $X = \bigcup_{j=1}^s X_{i,j}$ and for all $j,j'$ $X_{i,j} \cap X_{i,j'} = \emptyset$.  Let $\c{P}_X$ be the space of all partitions of $X$; since $X$ is fixed throughout this paper, we just refer to this space as $\c{P}$. 

There are two quantities that control the nature of the partitions generated. The \emph{quality} of a partition, represented by a function $Q : \c{P} \to \b{R}^+$, measures the degree to which a particular partition captures intrinsic structure in data; in general, most clustering algorithms that identify a single clustering attempt to optimize some notion of quality. The \emph{distance} between partitions, represented by the function $d : \c{P} \times \c{P} \to \b{R}$, is a quantity measuring how dissimilar two partitions are.  The partitions $\c{X}_i \in \c{P}$ that do a better job of capturing the structure of the data set $X$ will have a larger quality value $Q(\c{X}_i)$.  And the partitions $\c{X}_i, \c{X}_{i'} \in \c{P}$ that are more similar to each other will have a smaller distance value $d(\c{X}_i, \c{X}_{i'})$.  A good set of diverse partitions all have large distances from each other and all have high quality scores.  

Thus, the goal is this paper is to \emph{generate a set of $k$ partitions that best represent all high-quality partitions as accurately as possible}.  

\paragraph{Related Work.} 
There are two main approaches in the literature for computing many high-quality, diverse partitions.  However, both approaches focus only on a specific subproblem.  \emph{Alternate clustering} focuses on generating one additional partition of high-quality that should be far from a given set (typically of size one) of existing partitions.  \emph{$k$-consensus clustering} assumes an input set of many partitions, and then seeks to return $k$ representative partitions.  

Most algorithms for generating alternate partitions~\cite{Qi:2009:PFF:1557019.1557099,Davidson:2008:FAC:1510528.1511445,adco,Bae:2006:CNA:1193207.1193250,Dang:2010:HIT:1835804.1835878,Gondek:2004:NDC:1032649.1033439,DBLP:conf/sdm/DangB10} operate as follows. Generate a single partition using a clustering algorithm of choice. Next, find another partition that is both far from the first partition and of high quality. Most methods stop here, but a few methods try to discover more alternate partitions;  they repeatedly find new, still high-quality, partitions that are far from all existing partitions.  This effectively produces a variety of partitions, but the quality of each successive partition degrades quickly. 

Although there are a few other methods that try to discover alternate partitions simultaneously~\cite{metaclustering,Jain:2008:SUL:1465716.1465717,niu:multiple}, they are usually limited to discovering two partitions of the data. Other methods that generate more than just two partitions either randomly weigh the features or project the data onto different subspaces, but use the same clustering technique to get the alternate partitions in each round. Using the same clustering technique tends to generate partitions with clusters of similar shapes and might not be able to exploit all the structure in the data. 

The problem of $k$-consensus, which takes as input a set of $m \gg k$ partitions of a single data set to produce $k$ distinct partitions, has not been studied as extensively. To obtain the input for this approach, either the output of several distinct clustering algorithms, or the output of multiple runs of the same randomized algorithm with different initial seeds are considered~\cite{FLAIRS112624,ZhangLi}.
This problem can then be viewed as a clustering problem; that is, finding $k$ clusters of partitions from the set of input partitions. Therefore, there are many possible optimization criteria or algorithms that could be explored for this problem as there are for clustering in general. Most formal optimization problems are intractable to solve exactly, making heuristics the only option.  
Furthermore, no matter the technique, the solution is only as good as the input set of partitions, independent of the optimization objective. In most $k$-consensus approaches, the set of input partitions is usually not diverse enough to give a good solution.

In both cases, these subproblems avoid the full objective of constructing a diverse set of partitions that represent the \emph{landscape of all high-quality partitions}.  
The alternate clustering approach is often too reliant on the initial partition, has had only limited success in generalizing the initial step to generate $k$ partitions.
The $k$-consensus partitioning approach does not verify that its input represents the space of all high-quality partitions, so a representative set of those input partitions is not necessarily a representative set of all high-quality partitions.

\begin{figure}[h]
\centering
\includegraphics[width=0.7\linewidth]{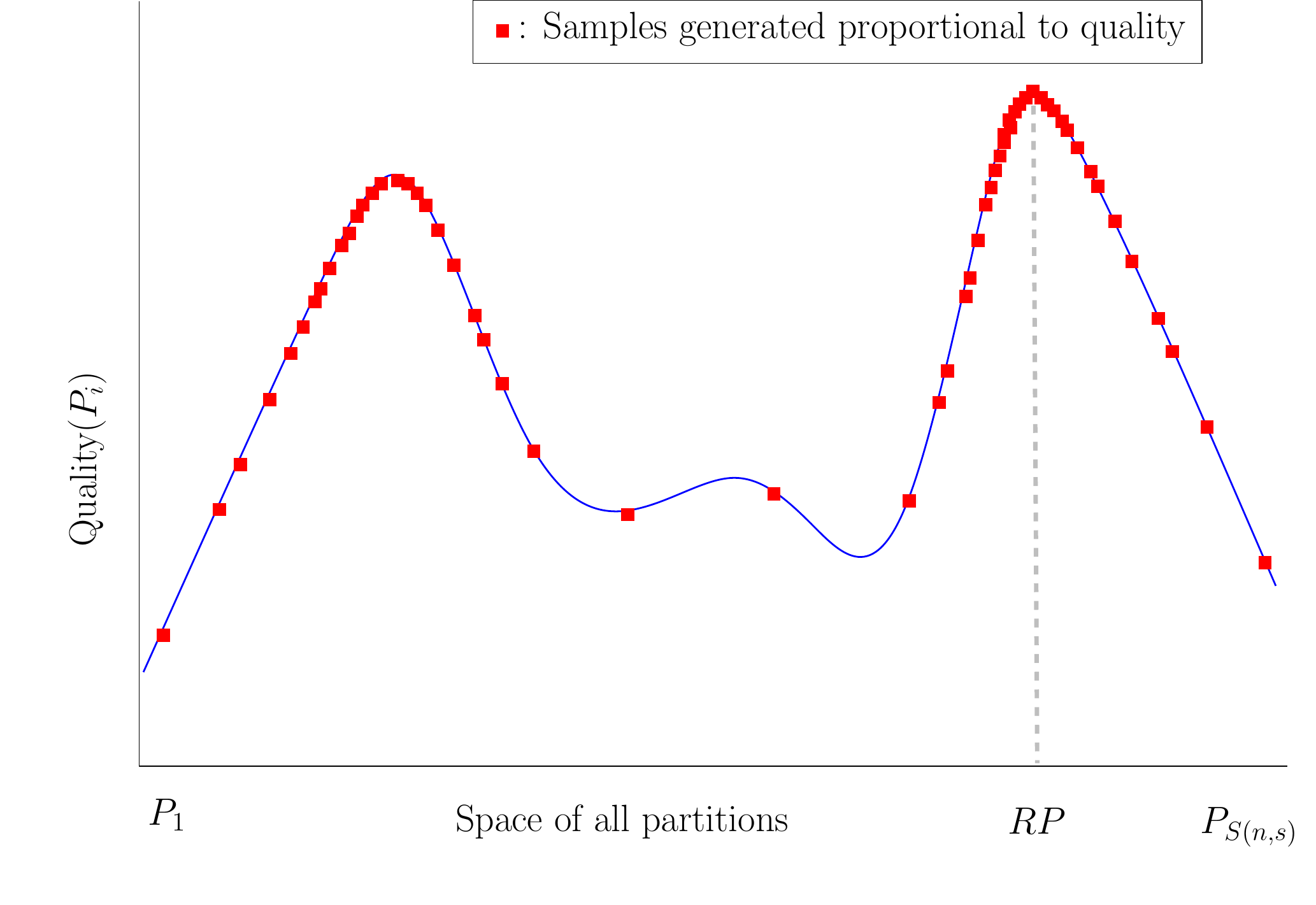}
\caption{\sffamily Sampling partitions proportional to its quality from the space of all partitions with $s$ clusters.\label{fig:sampling}}
\vspace*{-5.0ex}
\end{figure}

\paragraph{Our approach.}
To generate multiple good partitions, we present a new paradigm which decouples the notion of distance between partitions and the quality of partitions. Prior methods that generate multiple diverse partitions cannot explore the space of partitions entirely since the distance component in their objective functions biases against partitions close to the previously generated ones. These could be interesting partitions that might now be left out. To avoid this, we will first look at the space of all partitions more thoroughly and then pick non-redundant partitions from this set.
Let $k$ be the number of diverse partitions that we seek. Our approach works in two steps. 
In the first step called the \emph{generation step}, we first sample from the space of all partitions proportional to their quality. Stirling numbers of the second kind, $S(n,s)$ is the number of ways of partitioning a set of $n$ elements into $s$ nonempty subsets. Therefore, this is the size of the space that we sample from. We illustrate the sampling in figure \ref{fig:sampling}. This generates a set of size $m \gg k$ to ensure we get a diverse sample that represents the space of all partitions well, since generating only $k$ partitions in this phase may ``accidentally'' miss some high quality region of $\c{P}$. 
Next, in the \emph{grouping step}, we cluster this set of $m$ partitions into $k$ sets, resulting in $k$ clusters of partitions.  We then return one representative from each of these $k$ clusters as our output alternate partitions.  

Note that because the \emph{generation step} is decoupled from the \emph{grouping step}, we treat all partitions fairly, independent of how far they are from the existing partitions.  This allows us to explore the true density of high quality partitions in $\c{P}$ without interference from the choice of initial partition.
Thus, if there is a dense set of close interesting partitions our approach will recognize that. Also because the \emph{grouping step} is run separate from the \emph{generation step}, we can abstract this problem to a generic clustering problem, and we can choose one of many approaches. This allows us to capture different properties of the diversity of partitions, for instance, either guided just by the spatial distance between partitions, or also by a density-based distance which only takes into account the number of high-quality partitions assigned to a cluster. 

From our experimental evaluation, we note that decoupling the generation step from the grouping step helps as we are able to generate a lot of very high quality partitions. In fact, the quality of some of the generated partitions is better than the quality of the partition obtained by a consensus clustering technique called LiftSSD~\cite{liftemd}. The relative quality w.r.t. the reference partition of a few generated partitions even reach close to $1$. To our best knowledge, such partitions have not been uncovered by other previous meta-clustering techniques. The grouping step also picks out representative partitions far-away from each other. We observe this by computing the closest-pair distance between representatives and comparing it against the distance values of the partitions to their closest representative.

\paragraph{Outline.}
In Section \ref{sec:generation}, we discuss a sampling-based approach for generating many partitions proportional to their quality; i.e. the higher the quality of a partition, the more likely it is to be sampled. 
In Section \ref{sec:grouping}, we describe how to choose $k$ representative partitions from the large collection partitions already generated.  
We will present the results of our approach in Section \ref{sec:experiments}.  We have tested our algorithms on a synthetic dataset, a standard clustering dataset from the UCI repository and a subset of images from the Yale Face database B.


\section{Generating Many High Quality Partitions}
\label{sec:generation}

In this section we describe how to generate many high quality partitions.  This requires 
\begin{inparaenum}[(1)]
\item a measure of quality, and
\item an algorithm that generates a partition with probability proportional to its quality.
\end{inparaenum} 

\subsection{Quality of Partitions}

Most work on clustering validity criteria look at a combination of how compact clusters are and how separated two clusters are. Some of the popular measures that follow this theme are \emph{S}$\_$\emph{Dbw}, \emph{CDbw}, \emph{SD} validity index, maximum likelihood and Dunn index~\cite{989517,Halkidi:2000:QSA:645804.669820,springerlink:10.1007/BF02294245,patternrecognition,Dave:1996:VFP:230690.230704,MR1617519,MacKay:2002:ITI:971143,dunn74index}. Ackerman et. al. also discuss similar notions of quality, namely \emph{VR} (variance ratio) and \emph{WPR} (worst pair ratio) in their study of clusterability~\cite{citeulike:3680075,DBLP:journals/jmlr/AckermanB09}. We briefly describe a few specific notions of quality below.

\paragraph{$k$-Means quality.}
If the elements $x \in X$ belong to a metric space with an underlying distance $\delta : X \times X \to \b{R}$ and each cluster $X_{i,j}$ in a partition $\c{X}_i$ is represented by a single element $\bar x_j$, then we can measure the inverse quality of a cluster by 
$\bar q(X_{i,j}) = \sum_{x \in X_{i,j}} \delta(x, \bar x_j)^2$.  
Then the quality of the entire partition is then the inverse of the sum of the inverse qualities of the individual clusters: 
$\bar Q(\c{X}_i) = 1/(\sum_{j=1}^s \bar q(X_{i,j}))$.  

This corresponds to the quality optimized by $s$-mean clustering \footnote{it is commonplace to use $k$ in place of $s$, but we reserve $k$ for other notions in this paper}, and is quite popular, but is susceptible to outliers.  If all but one element of $X$ fit neatly in $s$ clusters, but the one remaining point is far away, then this one point dominates the cost of the clustering, even if it is effectively noise.  Specifically, the quality score of this measure is dominated by the points which fit least well in the clusters, as opposed to the points which are best representative of the true data. Hence, this quality measure may not paint an accurate picture about the partition.

\paragraph{Kernel distance quality.}
We introduce a method to compute quality of a partition, based on the kernel distance~\cite{kerneldistance}. Here we start with a similarity function between two elements of $X$, typically in the form of a (positive definite) kernel:  $K : X \times X \to \b{R}^+$.  If $x_1, x_2 \in X$ are more similar, then $K(x_1,x_2)$ is smaller than if they are less similar.  Then the overall similarity score between two clusters $X_{i,j}, X_{i,j'} \in \c{X}_i$ is defined 
$\kappa(X_{i,j},X_{i,j'}) = \sum_{x \in X_{i,j}} \sum_{x' \in X_{i,j'}} K(x,x')$, 
and a single clusters self-similarity for $X_{i,j} \in \c{X}_i$ is defined $\kappa(X_{i,j},X_{i,j})$.  
Finally, the overall quality of a partition is defined 
$Q_K(\c{X}_i) = \sum_{j=1}^s \kappa(X_{i,j},X_{i,j})$.

If $X$ is a metric space, the highest quality partitions divide $X$ into $s$ Voronoi cells around $s$ points -- similar to $s$-means clustering. However, its score is dominated by the points which are a good fit to a cluster, rather than outlier points which do not fit well in any cluster.  This is a consequence of how kernels like the Gaussian kernel taper off with distance, and is the reason we recommend this measure of cluster quality in our experiments.  

\subsection{Generation of Partitions Proportional to Quality}
We now discuss how to generate a sample of partitions proportional to their quality.  This procedure will be independent of the measure of quality used, so we will generically let $Q(\c{X}_i)$ denote the quality of a partition. Now the problem becomes to generate a set $Y \subset \c{P}$ of partitions where each $\c{X}_i \in Y$ is drawn randomly proportionally to $Q(\c{X}_i)$.  

The standard tool for this problem framework is a Metropolis-Hastings random-walk sampling procedure~\cite{Metropolis53,Hastings70,hoff09}.  Given a domain $X$ to be sampled and an energy function $Q : X \to \b{R}$, we start with a point $x \in X$, and suggest a new point $x_1$ that is typically ``near'' $x$. The point $x_1$ is accepted unconditionally if $Q(x_1) \ge Q(x)$, and is accepted with probability $Q(x_1)/Q(x)$ if not. Otherwise, we say that $x_1$ was \emph{rejected} and instead set $x_1 = x$ as the current state. After some sufficiently large number of such steps $t$, the expected state of $x_t$ is a random draw from $\c{P}$ with probability proportional to $Q$.  To generate many random samples from $\c{P}$ this procedure is repeated many times. 

In general, Metropolis-Hastings sampling suffers from high autocorrelation, where consecutive samples are too close to each other. This can happen when far away samples are rejected with high probability. To counteract this problem, often Gibbs sampling  is used~\cite{RGG97}.  Here, each proposed step is decomposed into several orthogonal suggested steps and each is individually accepted or rejected in order. This effectively constructs one longer step with a much higher probability of acceptance since each individual step is accepted or rejected independently. Furthermore, if each step is randomly made proportional to $Q$, then we can always accept the suggested step, which reduces the rejection rate. 

\paragraph{Metropolis-Hastings-Gibbs sampling for partitions.} 

The Metropolis-Hastings procedure for partitions works as follows.  Given a partition $\c{X}_i$, we wish to select a random subset  $Y \subset X$ and randomly reassign the elements of $Y$ to different clusters. If the size of $Y$ is large, this will have a high probability of rejection, but if $Y$ is small, then the consecutive clusters will be very similar. 
Thus, we use a Gibbs-sampling approach. At each step we choose a random ordering $\sigma$ of the elements of $X$. Now, we start with the current partition $\c{X}_i$ and choose the first element $x_{\sigma(1)} \in X$.  We assign $x_{\sigma(1)}$ to each of the $s$ clusters generating $s$ suggested partitions $\c{X}_i^j$ and calculate $s$ quality scores $q_j = Q(\c{X}_i^j)$.  Finally, we select index $j$ with probability $q_j$, and assign $x_{\sigma(1)}$ to cluster $j$. Rename the new partition as $\c{X}_i$. We repeat this for all points in order.  Finally, after all elements have been reassigned, we set $\c{X}_{i+1}$ to be the resulting partition. 

Note that auto-correlation effects may still occur since we tend to have partitions with high quality, but this effect will be much reduced.  Note that we do not have to run this entire procedure each time we need a new random sample. It is common in practice to run this procedure for some number $t_0$ (typically $t_0 = 1000$) of \emph{burn-in} steps, and then use the next $m$ steps as $m$ random samples from $\c{P}$. The rationale is that after the burn-in period, the induced Markov chain is expected to have mixed, and so each new step would yield a random sample from the stationary distribution. 

\section{Grouping the Partitions}
\label{sec:grouping}

Having generated a large collection $Z$ of $m \gg k$ high-quality partitions from $\c{P}$ by random sampling, we now describe a grouping procedure that returns $k$ representative partitions from this collection. 
We will start by placing a metric structure on $\c{P}$. This allows us to view the problem of grouping as a metric clustering problem. Our approach is independent of any particular choice of metric; obviously, the specific choice of distance metric and clustering algorithm will affect the properties of the output set we generate. 
There are many different approaches to comparing partitions. While our approach is independent of the particular choice of distance measure used, we review the main classes. 

\paragraph{Membership-based distances.}
The most commonly used class of distances is membership-based. These distances compute statistics about the number of \emph{pairs} of points which are placed in the same or different cluster in both partitions, and return a distance based on these statistics. Common examples include the Rand distance, the variation of information, and the normalized mutual information\cite{vi,Ran71,nmi,jaccard}. While these distances are quite popular, they ignore information about the spatial distribution of points within clusters, and so are unable to differentiate between partitions that might be significantly different. 

\paragraph{Spatially-sensitive distances.}
In order to rectify this problem, a number of \emph{spatially-aware} measures have been proposed. In general, they work by computing a concise representation of each cluster and then use the earthmover's distance (EMD)\cite{emd} to compare these sets of representatives in a spatially-aware manner. These include CDistance\cite{icml10}, $d_{\textsf{ADCO}}$\cite{adco}, CC distance\cite{zhou}, and LiftEMD\cite{liftemd}. As discussed in\cite{liftemd}, LiftEMD has the benefit of being both efficient as well as a well-founded metric, and is the method used here. 

\paragraph{Density-based distances.}
The partitions we consider are generated via a sampling process that samples more densely in high-quality regions of the space of partitions. 
In order to take into account dense samples in a small region, we use a \emph{density-sensitive} distance that intuitively spreads out regions of high density. 
Consider two partitions $\c{X}_i$ and $\c{X}_{i'}$.  Let $d : \c{P} \times \c{P} \to \b{R}^+$ be any of the above natural distances on $\c{P}$.  Then let $d_Z : \c{P} \times \c{P} \to \b{R}^+$ be a density-based distance defined as
$d_Z(\c{X}_i,\c{X}_{i'}) = | \{\c{X}_{l} \in Z \mid d(\c{X}_i, \c{X}_l) < d(\c{X}_i,\c{X}_{i'})\} |.$

\subsection{Clusters of Partitions}
Once we have specified a distance measure to compare partitions, we can cluster them. We will use the notation $\phi(\c{X}_i)$ to denote the representative partition $\c{X_i}$ is assigned to.  
We would like to pick $k$ representative partitions, and a simple algorithm by Gonzalez\cite{gonzalez} provides a $2$-approximation to the best clustering that minimizes the \emph{maximum} distance between a point and its assigned center. The algorithm maintains a set of centers $k' < k$ in $C$.  Let $\phi_C(\c{X}_i)$ represent the partition in $C$ closest to $\c{X}_i$ (when apparent we use just $\phi(\c{X}_i)$ in place of $\phi_C(\c{X}_i)$).  The algorithm chooses $\c{X}_i \in Z$ with maximum value $d(\c{X}_i, \phi(\c{X}_i))$.  It adds this partition $\c{X}_i$ to $C$ and repeats until $C$ contains $k$ partitions.  
We run the Gonzalez method to compute $k$ representative partitions using LiftEMD between partitions. We also ran the method using the density based distance derived from using LiftEMD. We got very similar results in both cases and we will only report the results from using LiftEMD in section \ref{sec:experiments}. We note that other clustering methods such as $k$-means and hierarchical agglomerative clustering yield similar results.

\section{Experimental Evaluation}
\label{sec:experiments}
In this section, we show the effectiveness of our technique in generating partitions of good divergence and its power to find partitions with very high quality, well beyond usual consensus techniques.

\paragraph{Data.} We created a synthetic dataset \textit{2D5C} with $100$ points in 2-dimensions, for which the data is drawn from $5$ Gaussians to produce 5 visibly separate clusters.
We also test our methods on the Iris dataset containing 150 points in 4 dimensions from UCI machine learning repository~\cite{uci}. We also use a subset of the Yale Face Database B~\cite{GeBeKr01} (90 images corresponding to 10 persons and 9 poses in the same illumination). The images are scaled down to 30x40 pixels. 

\paragraph{Methodology.} For each dataset, we first run $k$-means to get the first partition with the same number of clusters specified by the reference partition. Using this as a seed, we generate $m = 4000$ partitions after throwing away the first $1000$ of them. We then run the Gonzalez $k$-center method to find $10$ representative partitions. We associate each of the $3990$ remaining partitions with the closest representative partition. We compute and report the quality of each of these representative partitions. We also measure the LiftEMD distance to each of these partitions from the reference partition. For comparison, we also plot the quality of consensus partitions generated by LiftSSD~\cite{liftemd} using inputs from $k$-means, single-linkage, average-linkage, complete-linkage and Ward's method. 

\subsection{Performance Evaluation} 

\begin{figure}[h]
\centering
\vspace*{-5.0ex}
\subfigure{\includegraphics[width=0.6\linewidth]{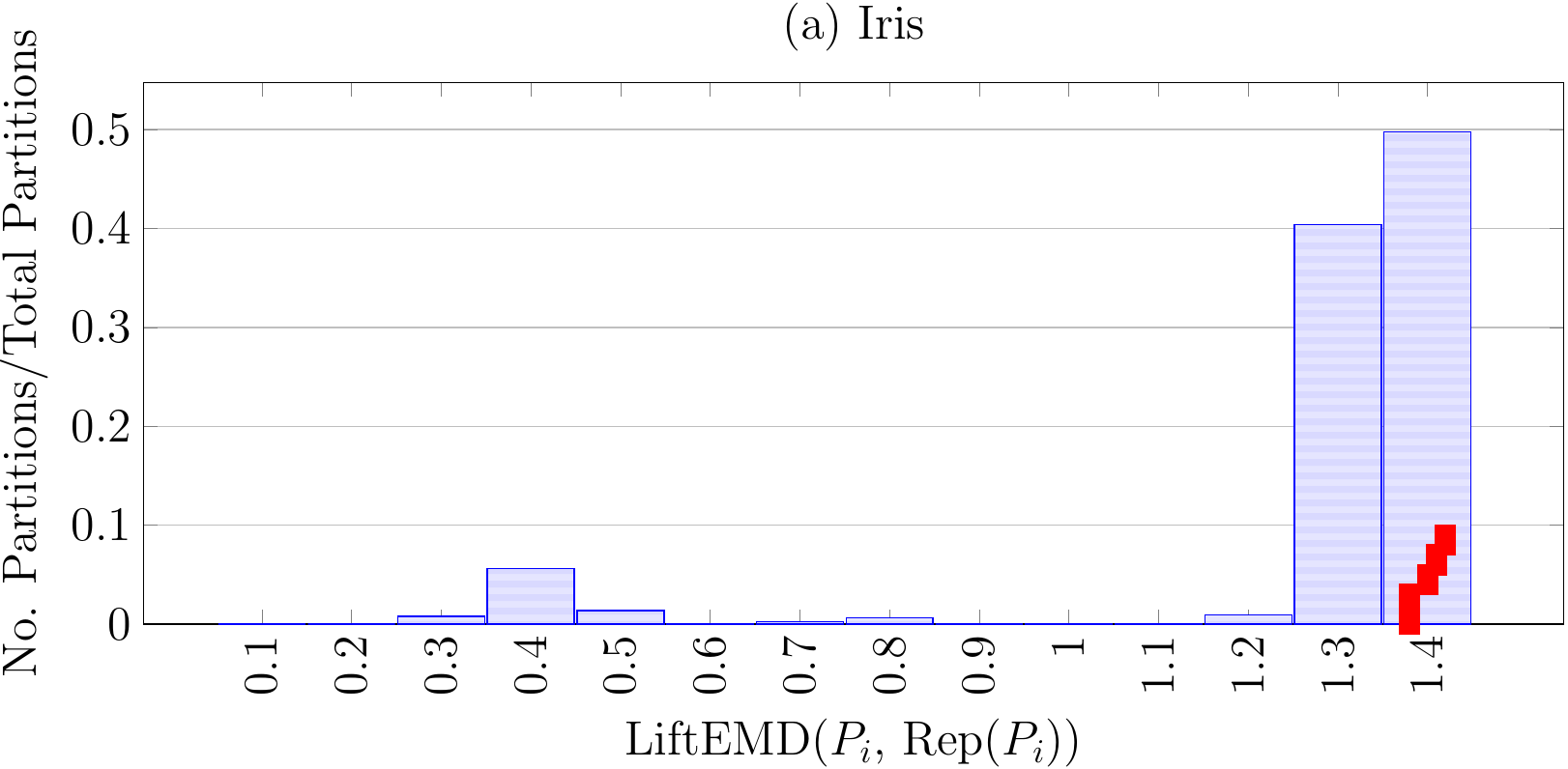}\label{fig:iris1}}
\subfigure{\includegraphics[width=0.6\linewidth]{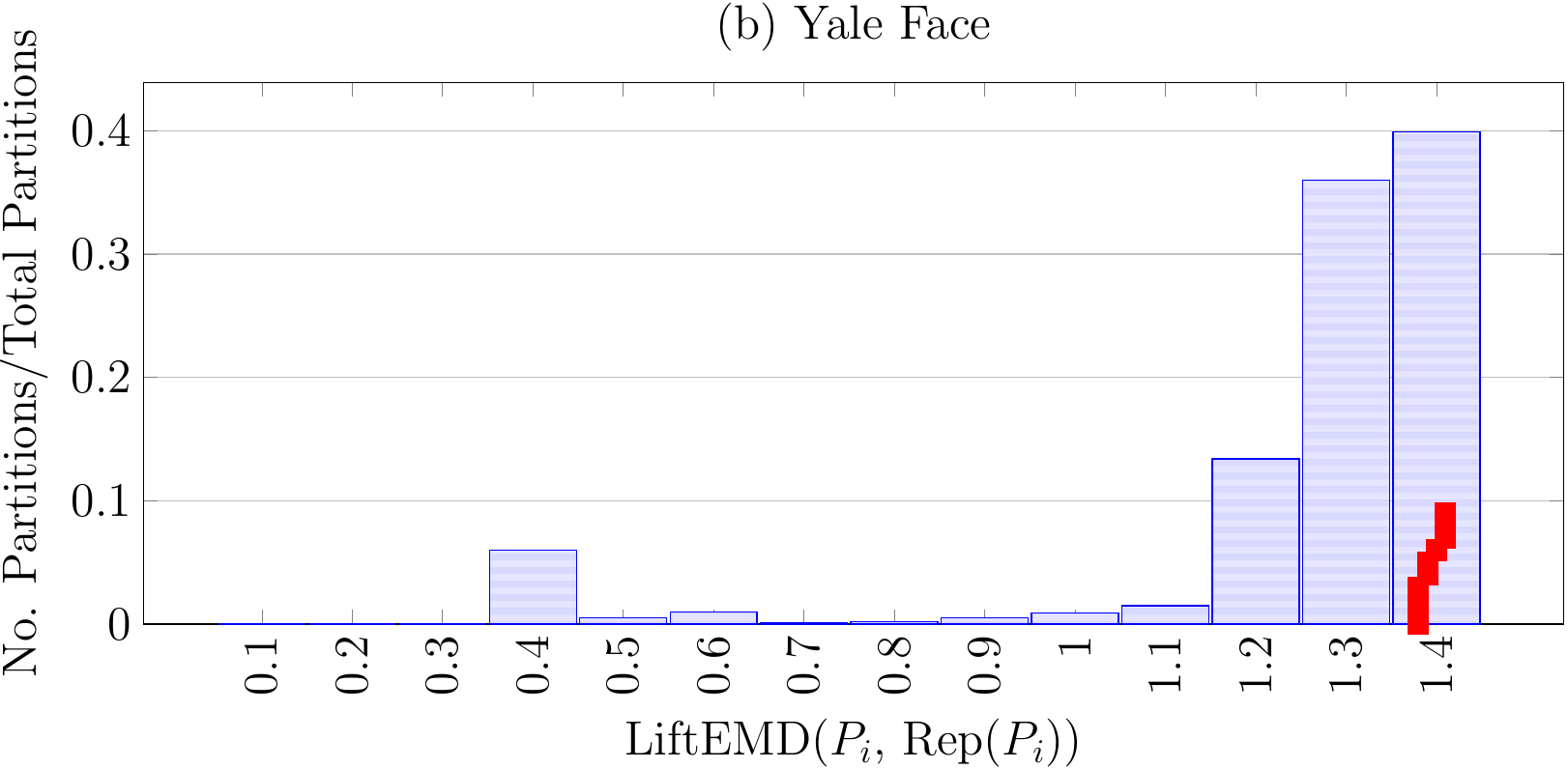}\label{fig:face1}}
\subfigure{\includegraphics[width=0.6\linewidth]{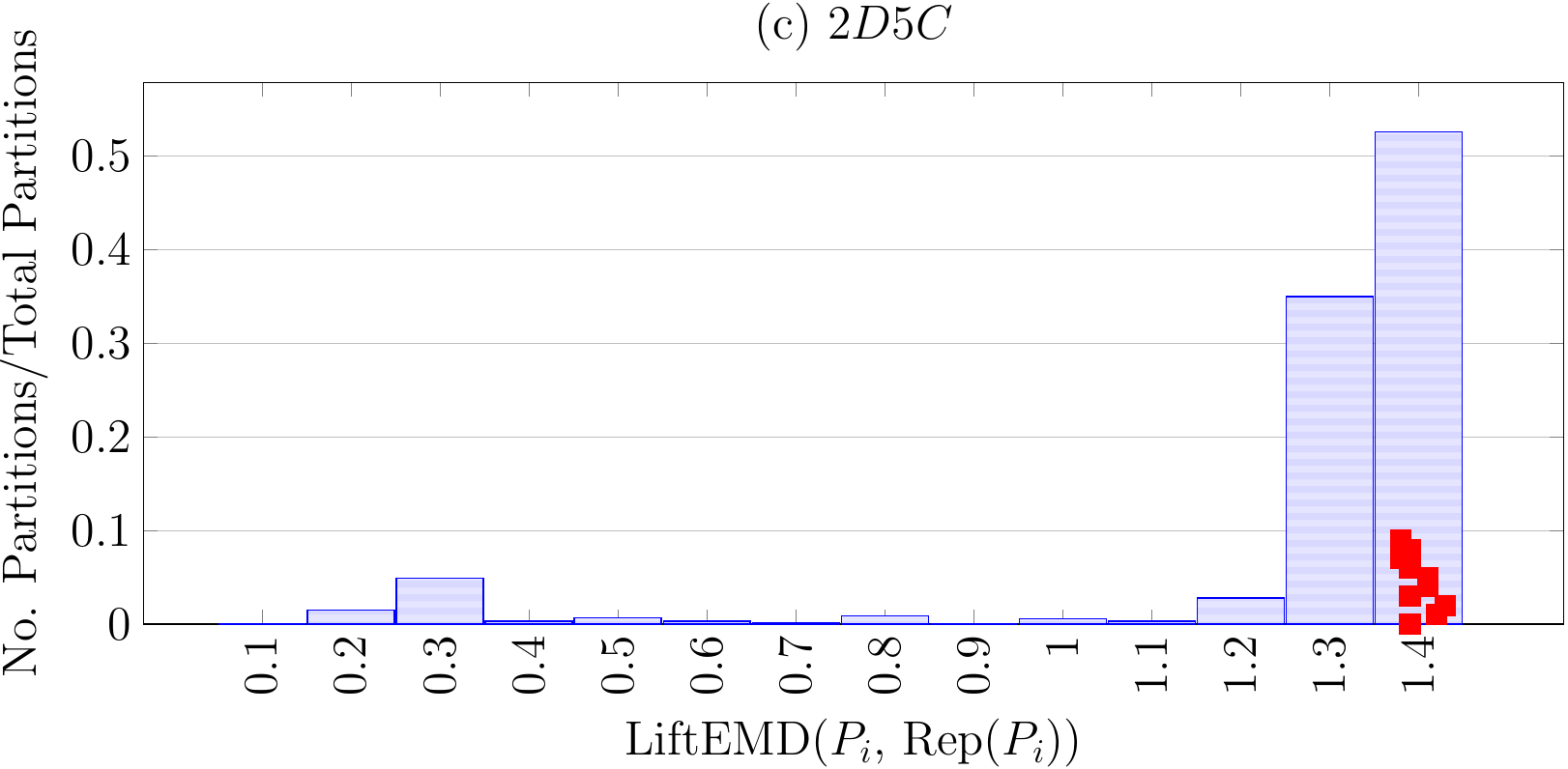}\label{fig:2d5c1}}
\caption{Distance between partition and its representative.} \label{distance}
\vspace*{-5.0ex}
\end{figure}

\paragraph{Evaluating partition diversity.}
We can evaluate partition diversity by determining how close partitions are to their chosen representatives using LiftEMD. Low LiftEMD values between partitions will indicate redundancy in the generated partitions and high LiftEMD values will indicate good partition diversity. The resulting distribution of distances is presented in Figures~\ref{fig:iris1},~\ref{fig:face1},~\ref{fig:2d5c1}, in which we also mark the distance values between a representative and its closest other representative with \emph{red} squares. Since we expect that the representative partitions will be far from each other, those distances provide a baseline for distances considered large. For all datasets, a majority of the partitions generated are generally far from the closest representative partition. For instance, in the Iris data set (\ref{fig:iris1}), about three-fourths of the partitions generated are far away from the closest representative with LiftEMD values ranging between $1.3$ and $1.4$.

\paragraph{Evaluating partition quality.}

\begin{figure}[t]
\centering
\vspace*{-5.0ex}
\subfigure{\includegraphics[width=0.6\linewidth]{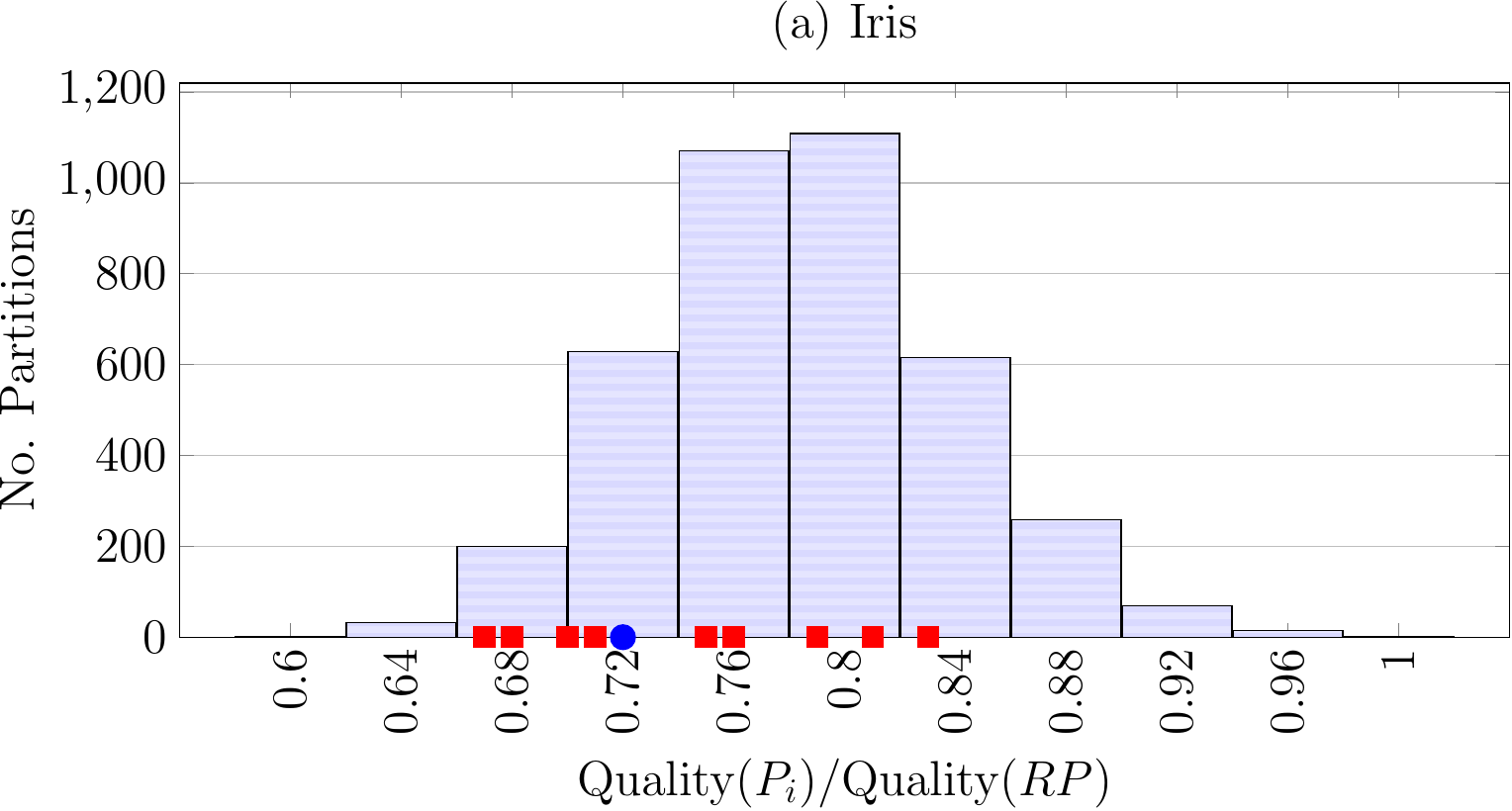}\label{fig:iris2}}
\subfigure{\includegraphics[width=0.6\linewidth]{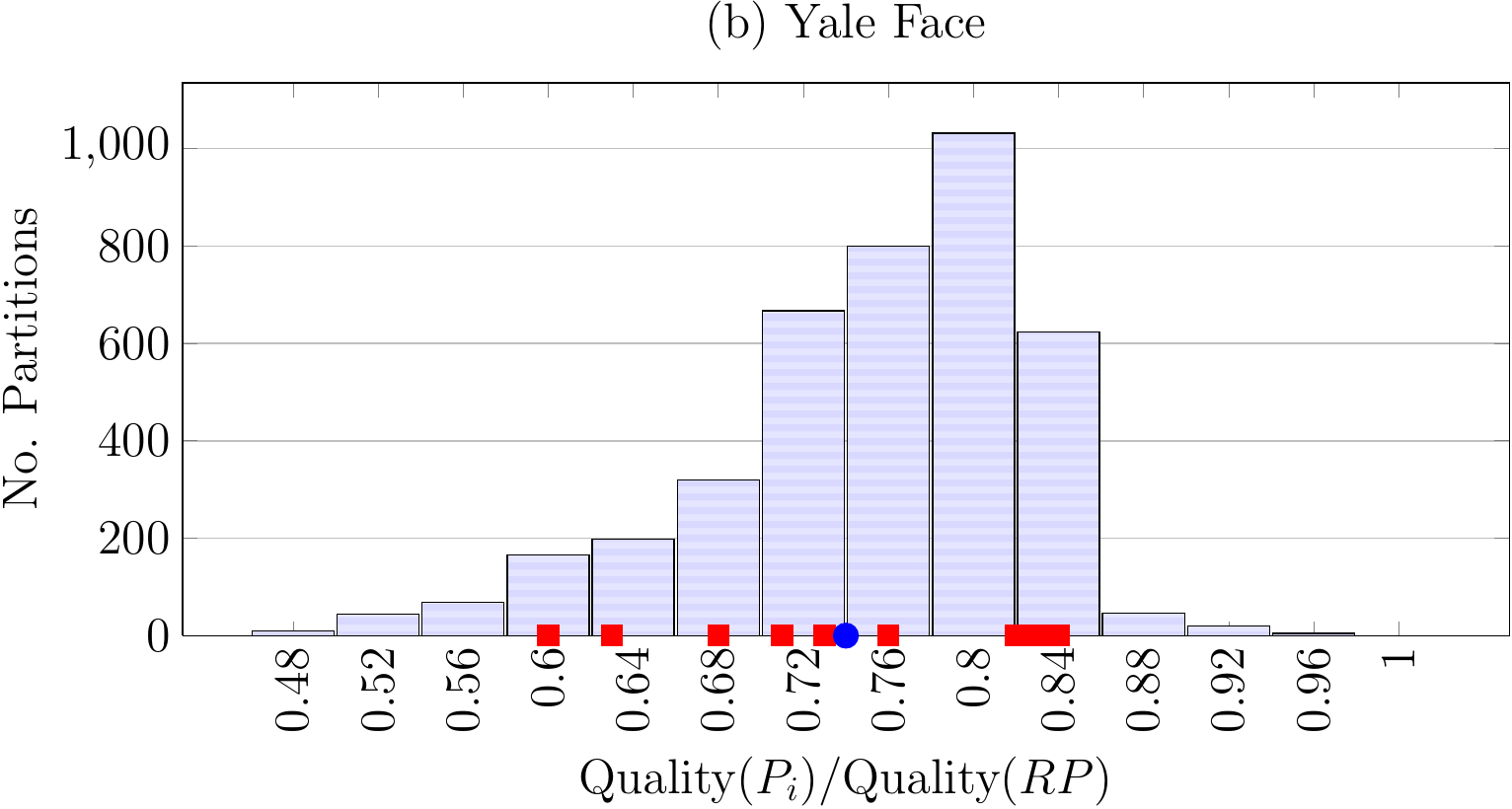}\label{fig:face2}}
\subfigure{\includegraphics[width=0.6\linewidth]{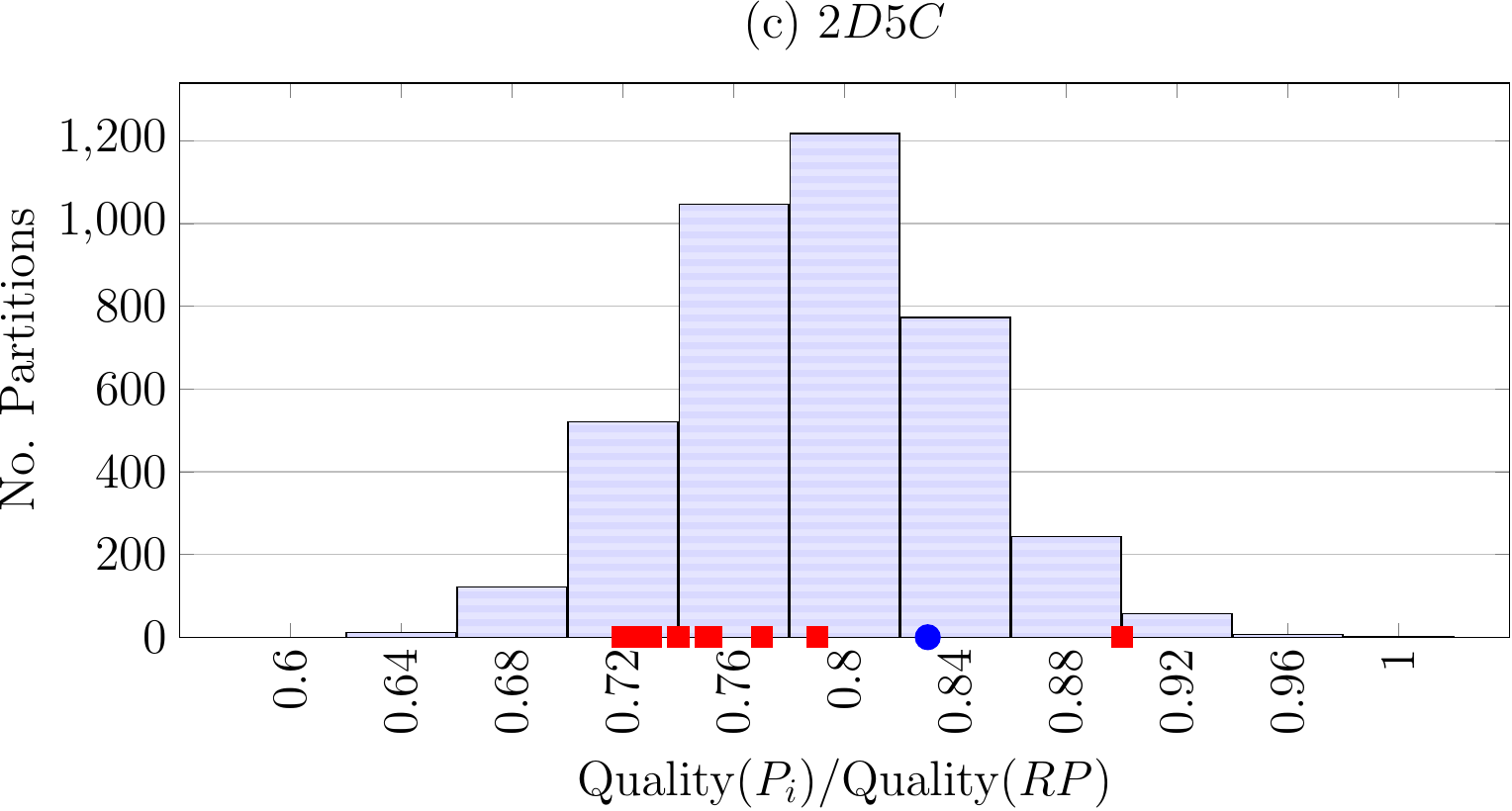}\label{fig:2d5c2}}
\caption{Quality of Partitions.} \label{quality}
\vspace*{-5.0ex}
\end{figure}

Secondly, we would like to inspect the quality of the partitions generated. Since we intend the generation process to sample from the space of all partitions proportional to the quality, we hope for a majority of the partitions to be of high quality. The ratio between the kernel distance quality $Q_K$ of a partition to that of the reference partition gives us a fair idea of the relative quality of that partition, with  values closer to 1 indicating partitions of higher quality. The distribution of quality is plotted in Figures~\ref{fig:iris2}\ref{fig:face2}\ref{fig:2d5c2}. 
We observe that for all the datasets, we get a normally distributed quality distribution with a mean value between $0.62$ and $0.8$. In addition, we compare the quality of our generated partitions against the consensus technique LiftSSD. We mark the quality of the representative partitions with \emph{red} squares and that of the consensus partition with a \emph{blue} circle. For instance, chart \ref{fig:iris2} shows that the relative quality w.r.t. the reference partition of three-fourths of the partitions is better than that of the consensus partition. For the Yale Face data, note that we have two reference partitions namely \textit{by pose} and \textit{by person} and we chose the partition \textit{by person} as the reference partition due to its superior quality.

\paragraph{Visual inspection of partitions.}

\begin{figure}[h]
\centering
\mbox{\subfigure[Iris]{\includegraphics[width=1.3in]{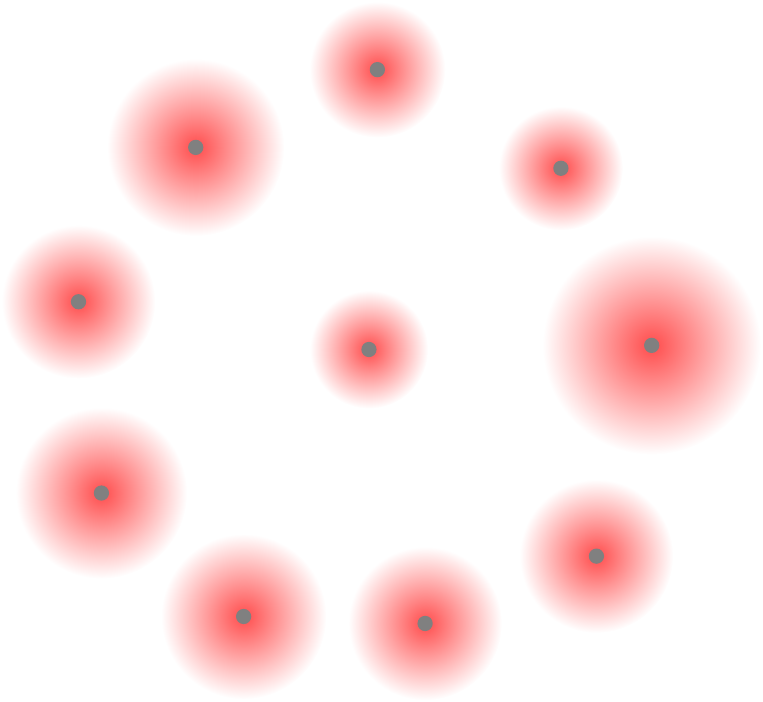}\label{fig:iris4}}\quad\hspace{12pt}
\subfigure[Yale Face]{\includegraphics[width=1.3in]{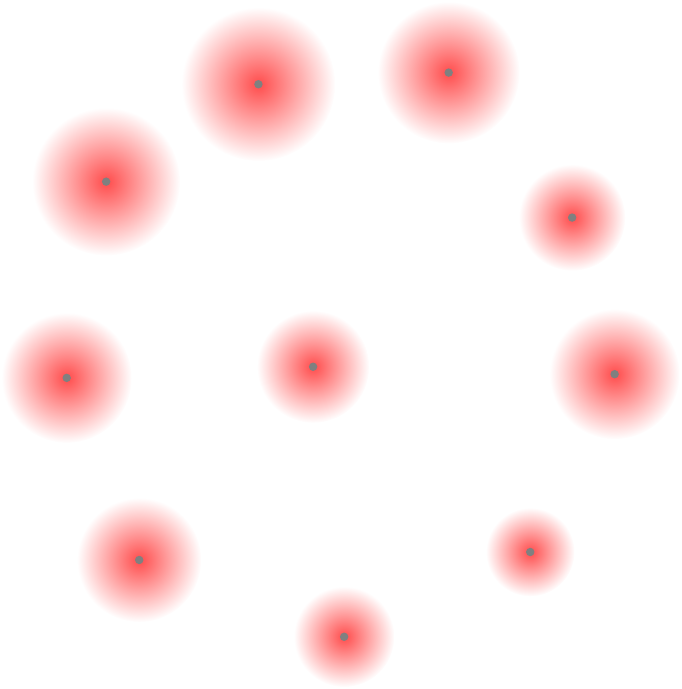}\label{fig:face4}}\quad\hspace{12pt}
\subfigure[$2D5C$]{\includegraphics[width=1.3in]{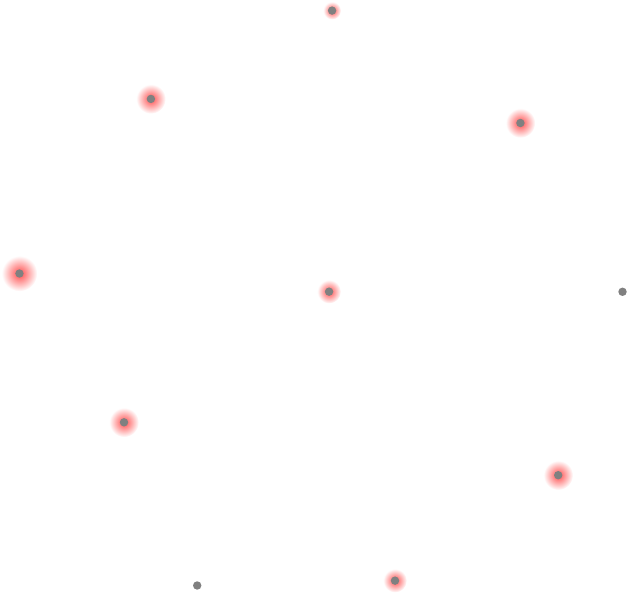}\label{fig:2d5c4}}}
\caption{MDS rendering of the LiftEMD distances between all representative partitions.} \label{mdsfig}
\end{figure}

\begin{figure}[h]
\centering
\subfigure{\includegraphics[width=0.9\linewidth]{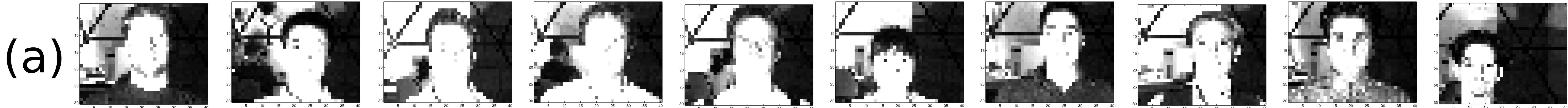}\label{fig:person}}
\subfigure{\includegraphics[width=0.9\linewidth]{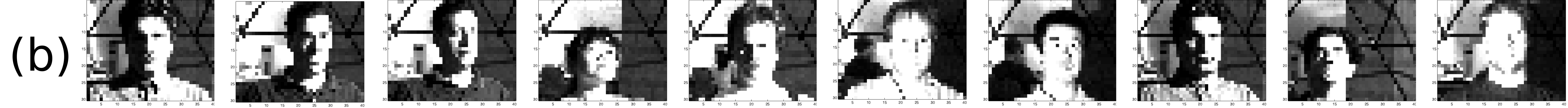}\label{fig:pose}}
\caption{Visual illustration of two interesting representative partitions on Yale Face.} \label{fig:faces}
\end{figure}

We ran multi-dimensional scaling\cite{mds} on the all-pairs distances between the $10$ representatives for a visual representation of the space of partitions. We compute the variance of the distances of the partitions associated with each representative and draw Gaussians around them to depict the \emph{size} of each cluster of partitions. For example, for the Iris dataset, as we can see from chart \ref{fig:iris4}, the clusters of partitions are well-separated and are far from the original reference partition. In figure \ref{fig:faces}, we show two interesting representative partitions on the Yale face database. We show the mean image from each of the 10 clusters. Figure \ref{fig:person} is a representative partition very similar to the partition \textit{by person} and figure \ref{fig:pose} resembles the partition \textit{by pose}.

\section{Conclusion}
In this paper we introduced a new framework to generate multiple non-redundant partitions of good quality. Our approach is a two stage process: in the \emph{generation} step, we focus on sampling a large number of partitions from the space of all partitions proportional to the quality and in the \emph{grouping} step, we identify $k$ representative partitions that best summarizes the space of all partitions. 

\begin{spacing}{0.9}
\bibliography{alter}
\bibliographystyle{abbrv}
\end{spacing}

\end{document}